\documentclass{isprs}
\usepackage{hyperref}
\hypersetup{
    colorlinks,
    filecolor=black,
    linkcolor=black,
    urlcolor=black
}
\usepackage{url}
\usepackage{subfigure}
\usepackage{setspace}
\usepackage{geometry} 
\usepackage{epstopdf}
\usepackage[labelsep=period]{caption}  
\usepackage{multirow}
\usepackage{amsmath}
\usepackage[standard,amsmath,thref]{ntheorem}
\usepackage{tikz}
\usetikzlibrary{arrows.meta}
\usepackage{nicefrac}

\usepackage{todonotes}

\usepackage{enumitem}

\begin{document}

\title{end-to-end depth from motion with stabilized monocular videos}

\author{
 C. Pinard\textsuperscript{a,b}\thanks{Corresponding author} , L. Chevalley\textsuperscript{a}, A. Manzanera\textsuperscript{b}, D. Filliat\textsuperscript{b}}

\address{
	\textsuperscript{a }Parrot, Paris, France - (clement.pinard, laure.chevalley)@parrot.com\\
	\textsuperscript{b }ENSTA, U2IS Lab, Palaiseau, France - (clement.pinard, antoine.manzanera, david.filliat)@ensta-paristech.fr
}


\commission{}{} 
\workinggroup{} 
\icwg{}   

\abstract{
We propose a depth map inference system from monocular videos based on a novel dataset for navigation that mimics aerial footage from gimbal stabilized monocular camera in rigid scenes. Unlike most navigation datasets, the lack of rotation implies an easier structure from motion problem which can be leveraged for different kinds of tasks such as depth inference and obstacle avoidance. We also propose an architecture for end-to-end depth inference with a fully convolutional network. Results show that although tied to camera inner parameters, the problem is locally solvable}


\maketitle

\section{Introduction}
Scene understanding from vision is a core problem for autonomous vehicles and for UAVs in particular.
In this paper we are specifically interested in computing the
depth of each pixel from a pair of consecutives images captured by a camera. We assume our camera's velocity (and thus movement between two frames) is known, as most UAV flight systems include a speed estimator, allowing to settle the scale invariance ambiguity.

Solving this problem could be beneficial for applying depth-based sense and avoid algorithms for lightweight embedded systems that only have a monocular camera and cannot directly provide an RGB-D image. This could allow such devices to go without heavy or power expensive dedicated devices such as ToF camera, LiDar or Infra Red emitter/receiver \cite{hitomi20153d} that would greatly lower autonomy. In addition, along with some being unable to operate under sunlight (e.g. IR and ToF), most RGBD sensor suffer from range limitations and can be inefficient in case we need long-range trajectory planning \cite{hadsell2009learning}. The faster an UAV is, the longer range we will need to efficiently avoid obstacles.
Unlike RGB-D sensors, depth from motion is robust to high speeds since it will be normalized by the displacement between two frames. Given the difficulty of the task, several learning approaches have been proposed to solve it.

A large number of datasets has been developed in order to propose supervised learning and validation for fundamental vision tasks, such as optical flow \cite{geiger2012we,DFIB15,weinzaepfel:hal-00873592} stereo disparity and even 3D scene flow \cite{menze2015object,MIFDB16}. These different measures can help figure up scene structure and camera motion, but they remain low-level in terms of abstraction.
End-to-end learning of a certain high semantic value such as three dimensional geometry may be hard to compute on a totally unrestricted monocular camera movement.

We focus on RGB-D datasets that would allow supervised learning of depth. RGB pairs (preferably with the corresponding displacement) being the input, and D the desired output.
Our choice today to learn depth from motion in existing RGB-D datasets is either unrestricted w.r.t. ego-motion \cite{firman-cvprw-2016,sturm12iros}, or a simple  stereo vision, equivalent to lateral movement \cite{geiger2012we,scharstein2002taxonomy}.

We thus propose a new dataset, described Part \ref{our_dataset}, which aims at proposing a bridge between the two by assuming that rotation is canceled on the footage that contains only random translations.

\begin{figure}
\begin{tabular}{rl}
a)\includegraphics[scale=0.1]{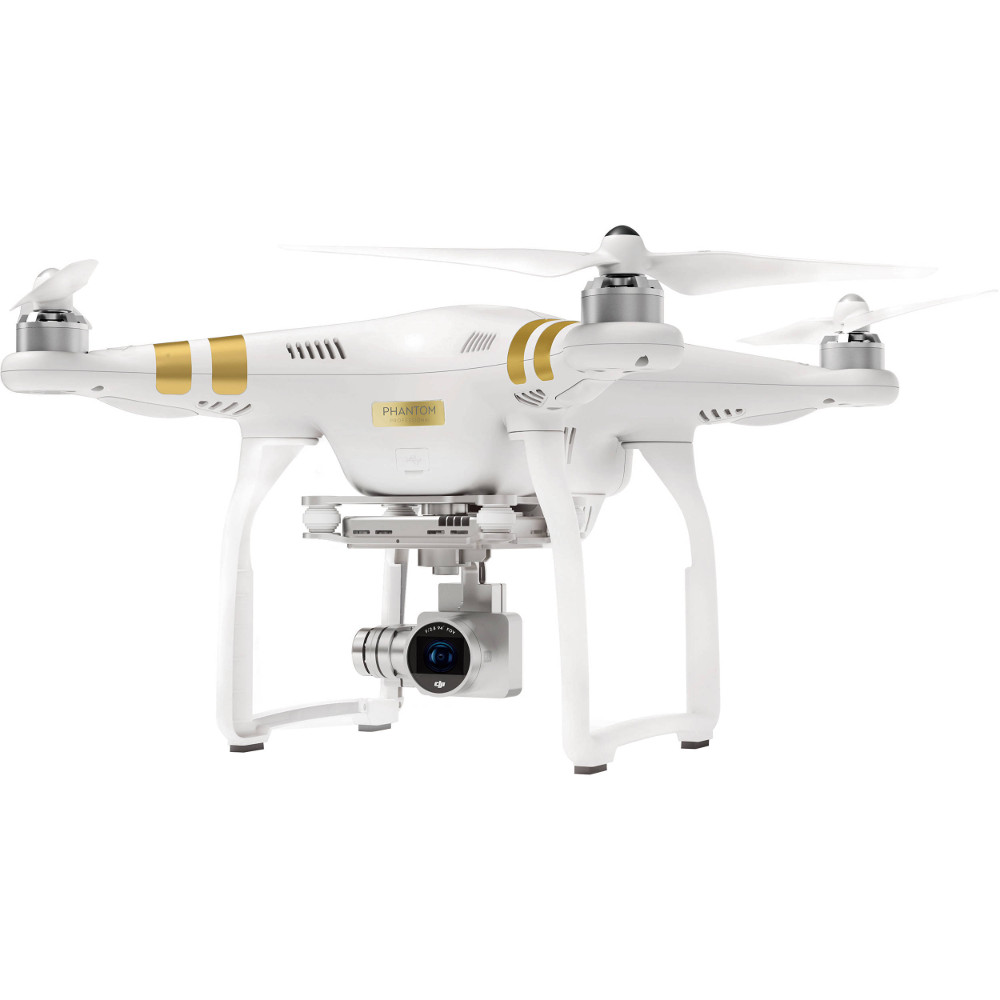} &
b)\includegraphics[scale=0.1]{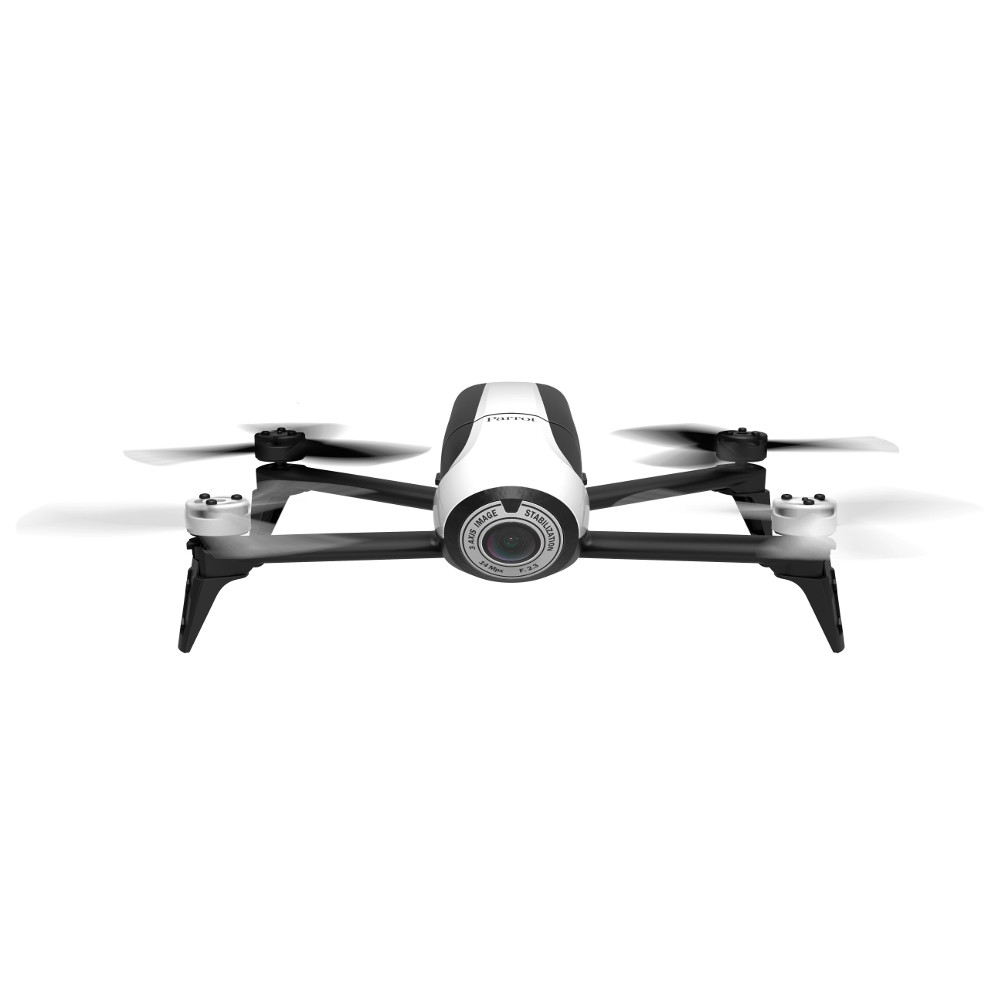} \\
\multicolumn{2}{c}{c)\includegraphics[scale=0.2]{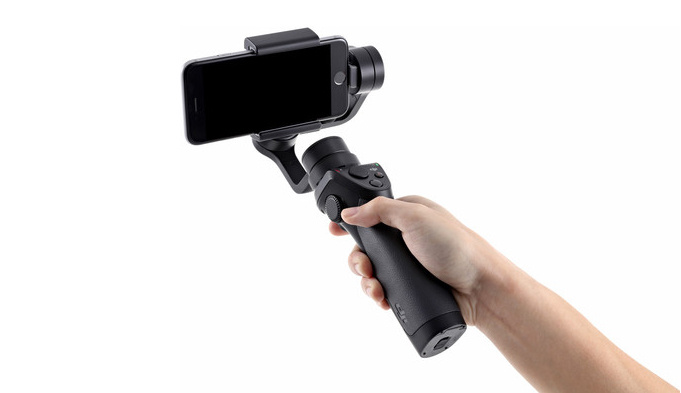}}\\
\end{tabular}
\caption{Camera stabilization can be done via a) mechanic gimbal or b) dynamic cropping from fish-eye camera, for drones or c) hand-held cameras}
\label{drones}
\end{figure}

This assumption about videos without rotation appears realistic for two reasons :
\begin{enumerate}
\setlength\itemsep{0em}\setlength\parskip{0em}\setlength\topsep{0em}\setlength\partopsep{0em}\setlength\parsep{0em}
\item Hardware rotation compensation is mainly a solved problem, even for consumer products, with IMU-stabilized cameras on consumer drones or hand-held steady-cam (Fig~\ref{drones}).
\item this movement is somewhat related to human vision and vestibulo-ocular reflex (VOR) \cite{VOR}. Our eyes orientation is not induced by head rotation, our inner ear among other biological sensors allows us to compensate parasite rotation when looking at a particular direction.
\end{enumerate}
This assumption allows to dramatically simplify links between optical flow and depth and leverage much simpler computation. The main benefit being the camera movement's dimensionality, reduced from 6 (translation and rotation) to 3 (only translation). However, as discussed in Part \ref{depth_part}, depth is not computed as simply as with stereo vision and requires being able to compute higher abstractions to avoid a possible indeterminate form, especially for forward movements.

Using the proposed dataset, we then show that depth can be learned as an end-to-end problem just like other usual Deep Learning problems. With a trained artificial neural network, we perform much better depth accuracy than flow based methods and are confident this will be efficiently leveraged for sense and avoid algorithms.

\section{Related Work}

\subsection{Monocular vision based sense and avoid}

Sense and avoid problems are mostly approached using a dedicated sensor for 3D analysis. However, some work has been done trying to leverage Optical flow from Monocular camera \cite{souhila2007optical,zingg2010mav}. These works enlighten the difficulty in estimating depth solely with flow, especially when the camera is pointed toward movement. One can note that rotation compensation was already used with fish-eye camera in order to have a more direct link between flow and depth. Another work \cite{coombs1998real} also demonstrated that basic obstacle avoidance could be achieved in cluttered environments such as a closed room.

Some interesting work concerning obstacle avoidance from Monocular camera \cite{lecun2005off,hadsell2009learning,michels2005high} showed that single frame analysis can be more efficient than depth from stereo for path planning. However, these works were not applied on UAV, on which depth cannot be directly deduced from distance to horizon, because obstacles and paths are now three-dimensional

More recently, Giusti et al. \cite{giusti2016machine} showed that a monocular system can be trained to follow a hiking path. But once again, only 2D movement is approached, asking a UAV going forward to change its yaw based on likeliness to be following a traced path.

\subsection{Depth inference}

Deep Learning and Convolutional Neural Networks has recently been widely used for numerous kinds of vision problem such as classification \cite{krizhevsky2012imagenet} and hand-written digits recognition \cite{lecun1998gradient}.

Depth from vision is one the problems studied with neural network, and has been addressed not only with image pairs, but also single images \cite{eigen2014depth,saxena2005learning}. Depth inference from stereo has also been widely studied \cite{luo2016efficient,zbontar2015computing}, and not necessarily in a supervised way \cite{DBLP:journals/corr/KondaM13,DBLP:journals/corr/GargBR16}.

Current state of the art methods for depth from monocular view tend to use motion, and especially structure from motion, and most algorithm do not rely on deep learning \cite{cadena2016past,mur2016orb,klein2007parallel}. Prior knowledge w.r.t. scene is used to infer a sparse depth map with its density usually growing over time. These techniques also called SLAM are typically used with unstructured movement, produce very sparse point-cloud based 3D maps and require heavy calculation to keep track of the scene structure and align newly detected 3D points to the existing ones. SLAM is not widely used for obstacle avoidance, but more for off-line 3D scan.

Our goal is to compute a dense (where every point has a valid depth) quality depth map using only two images, and without  prior knowledge on the scene and movement, apart from the lack of rotation and the scale factor.

\subsection{Navigation datasets}
As discussed earlier, numerous datasets exist with depth groundtruth, but to our knowledge, no dataset propose only translational movement. Some provide IMU data along with frames \cite{smith2009new}, that could be used to compensate rotation but their small size only allows us to use it as a validation set.

\section{Still Box Dataset}
\label{our_dataset}
\begin{table}
\centering
\begin{tabular}{|l|l|l|}
  \hline
  \multicolumn{3}{|c|}{Still Box Dataset} \\
  \hline
  image size & number of scenes & total size (GB) \\ \hline
  64x64 & 80K & 19 \\ \hline
  128x128 & 16K & 12 \\ \hline
  256x256 & 3.2K & 8.5 \\ \hline
  512x512 & 3.2K & 33 \\ \hline
\end{tabular}
\caption{datasets sizes}
\label{params1}
\end{table}
\begin{table}
\centering
\begin{tabular}{|r|l|}
 \hline
  \multicolumn{2}{|c|}{Scenes parameters} \\ \hline
  field of view & $90^o$ \\
  max render distance & $200m$ \\
  primitives number & $20$ \\
  texture ratio & $0.5$ \\
  size range of meshes (m) & $[0,2]$ \\
  distance range of meshes (m) & $[0,25]$ \\
  displacement & $10cm$ \\
  length (frames) & $10$ \\
  nominal shift & $3$ \\
  speed equivalent (for 30fps) & $9m.s^{-1}$ \\
   \hline
\end{tabular}
\caption{datasets parameters}
\label{params2}
\end{table}

\begin{figure}
\begin{tabular}{ll}
  \includegraphics[scale=0.2]{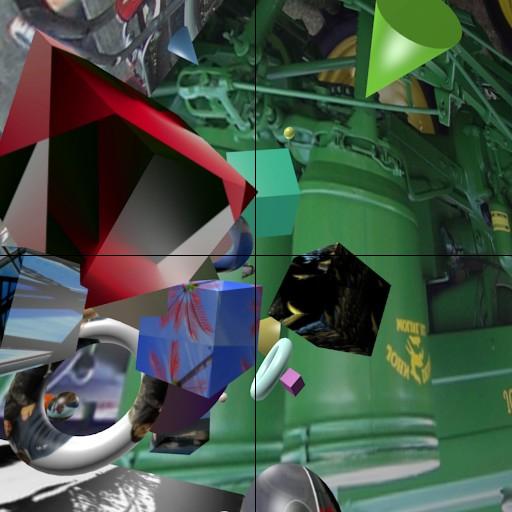} & \includegraphics[scale=0.2]{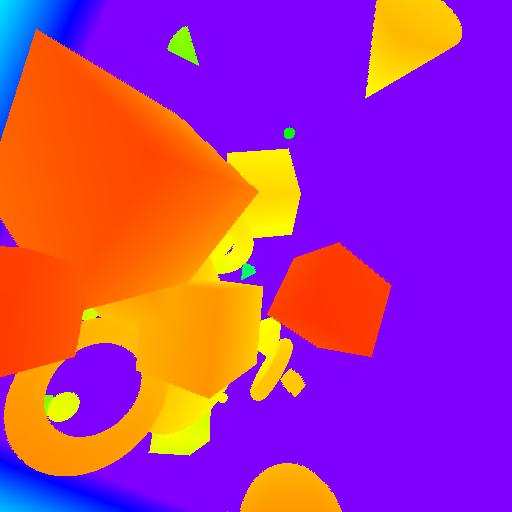} \\
  \includegraphics[scale=0.2]{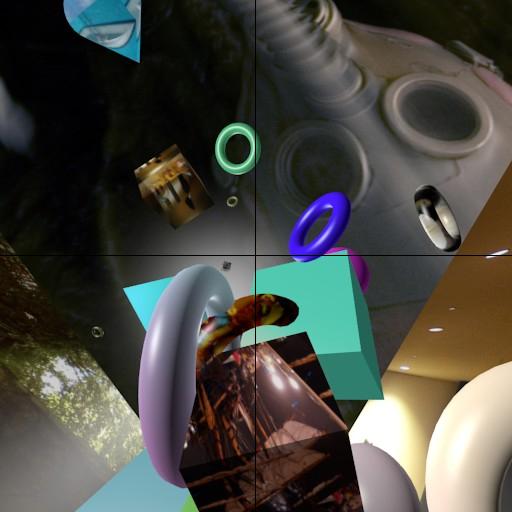} & \includegraphics[scale=0.2]{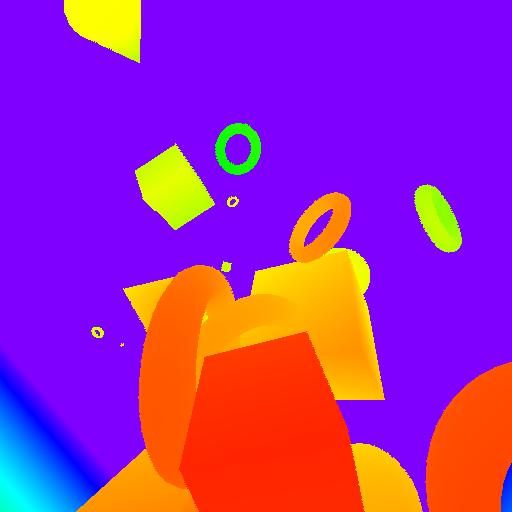} \\
  \includegraphics[scale=0.2]{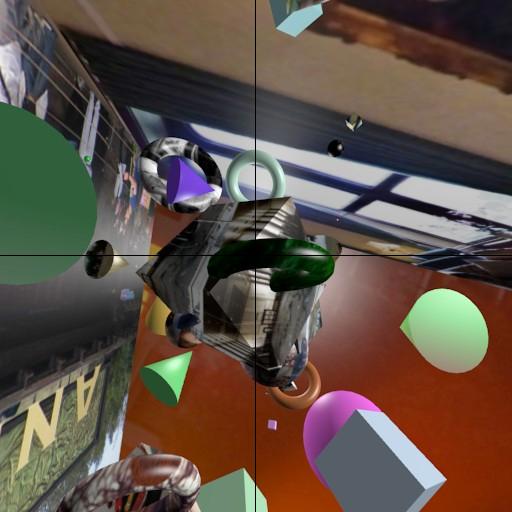} & \includegraphics[scale=0.2]{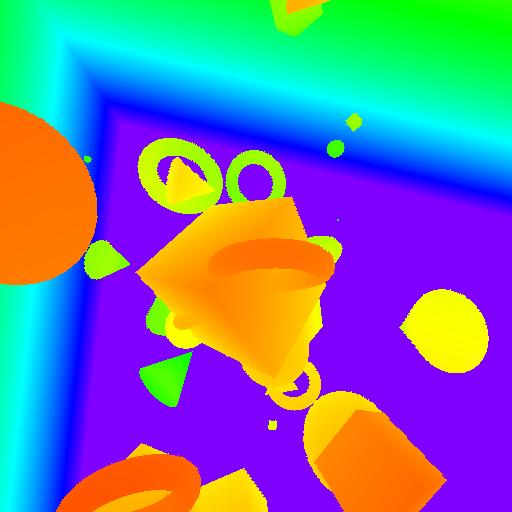}
\end{tabular}
\caption{Some examples of our renderings with associated depth maps (red is close, purple is far)}
\label{pictures}
\end{figure}

For our dataset we used the rendering software {\it Blender} to generate an arbitrary number of random rigid scenes, composed of basic 3d primitives (cubes, spheres, cones and tores) randomly textured from an image set scrapped from {\it Flickr} (see Fig~\ref{pictures}).

These objects are randomly placed and sized in the scene, so that they are mostly in front of the camera, with possible variations including objects behind camera, or even camera inside an object. Scenes in which camera goes through objects are discarded. To add difficulty we also applied uniform textures on a proportion or of the primitives. Each primitive thus has a uniform probability (corresponding to texture ratio) of being textured from a color-ramp and not from a photograph.

Walls are added at large distances as if the camera was inside a box (hence the name). The camera is moving at a fixed speed value, but to a random direction (uniform distribution), which is constant for each scene. It can be anything from forward/backward movement to lateral movement (which is then equivalent to stereo vision). Tables~\ref{params1} and \ref{params2} show a summary of our scenes parameters. They can be changed at will, and are stored in a metadata JSON file to keep track of it.
Our dataset is then composed of 4 sub-datasets with different resolutions, 64px dataset being the largest in term of number of samples, 512px being the heaviest in data.

\section{End-to-end learning of Depth Inference}
\label{depth_part}
\subsection{Why not disparity ?}
Flow Estimation and disparity (which is essentially magnitude of optical flow vectors) are problems to which exist a lot of very convincing methods \cite{ilg2016flownet,2017arXiv170304309K}.
Knowing depth and displacement in our dataset, we could be able to easily get disparity and train a network for it using existing methods. We consider a picture with $(u,v)$ coordinates, and optical center at  $\textbf{P}_0 = \begin{pmatrix}
u_0\\v_0\end{pmatrix}$

\begin{Definition}
Disparity is defined by the norm of a flow vector $\textbf{flow(P)} = \begin{pmatrix}
du\\dv\end{pmatrix}$ of a point $\textbf{P} = \begin{pmatrix}
u\\v\end{pmatrix}$.
\[
\forall \textbf{P} = \begin{pmatrix}
u\\v\end{pmatrix}, disparity(\textbf{P}) = \left\lVert\textbf{flow(P)}\right \rVert
\]
\end{Definition}

\begin{Definition}
Focus of Expansion is defined by the point $\textbf{FOE}$ where each flow vector $\textbf{flow(P)} = \begin{pmatrix}
du\\dv\end{pmatrix}$ of a point $\textbf{P} = \begin{pmatrix}
u\\v\end{pmatrix}$ is headed from. Note that this property is true only when considering no rotation and a rigid scene. One can note than for a pure translation, $\textbf{FOE}$ is the projection of the displacement vector
\[
\forall \textbf{P} = \begin{pmatrix}
u\\v\end{pmatrix}, \left\langle \overrightarrow{\textbf{P - FOE}} . \overrightarrow{\textbf{flow(P)}} \right\rangle= 0
\]
\end{Definition}

\begin{Theorem}
\label{D2D}
For a random rotation-less displacement of norm $V$ of a pinhole camera, with a focal length of $f$, depth is an explicit function of disparity ,focus of expansion $\textbf{FOE}$ and optical center $\textbf{P}_0$
\[
\forall \textbf{P},depth(\textbf{P}) = \frac{Vf}{\sqrt{f^2 + \left\lVert \textbf{P}_0 - \textbf{FOE} \right\rVert^2}} \left(\frac{\left\lVert \textbf{P} - \textbf{FOE} \right\rVert}{disparity(\textbf{P})}-1\right)
\]
\end{Theorem}

This result is in a useful form for limit values. Lateral movement corresponds to $\left\lVert \textbf{FOE}\right\rVert \to +\infty$ and then

\[ \lim_{\left\lVert \textbf{FOE}\right\rVert \to +\infty} depth(\textbf{P}) = \frac{fV}{disparity(\textbf{P})} \]

When approaching $\textbf{FOE}$, knowing depth is a bounded positive value, we can deduce :
\[
disparity(\textbf{P}) \underset{\textbf{P} \to \textbf{FOE}}{\varpropto}
\left\lVert \textbf{P} - \textbf{FOE} \right\rVert
\]

limit of disparity is this case is $0$ and we use its inverse. As a consequence, small errors on disparity estimation will result in diverging values of depth near focus of expansion while it corresponds to the direction the camera is moving to, which is clearly problematic for depth-based obstacle avoidance.

Given the random direction of our camera's displacement, computing depth from disparity is therefore much harder than for a classic stereo rig. To tackle this problem, we decided to set up an end-to-end learning workflow, by training a neural network to explicitly predict the depth of every pixel in the scene, from an image pair with constant displacement value $V$.

\subsection{Dataset set augmentation}

The way we store data in 10 images long videos, with each frame paired with its ground truth depth allows us to set {\it a posteriori} distances distribution with a variable temporal shift between two frames. 
If we use a baseline shift of 3 frames, we can e.g. assume a depth three times as great for two consecutive frames (shift of 1). 
In addition, we can also consider negative shift, which will only change displacement direction without changing speed value compared to opposite shift. This allows us, given a fixed dataset size, to get more evenly distributed depth values to learn, and also to de-correlate images from depth, preventing any over-fitting during training, 
that would result in a scene recognition algorithm and would perform poorly on a validation set.

\subsection{Depth Inference training}

\begin{figure}
\begin{tabular}{|l|}
  \hline
  Typical Conv Module \\ \hline \hline
  SpatialConv, 3x3\\ \hline
  SpatialBatchNorm\\ \hline
  ReLU\\ \hline
  
\end{tabular}
\begin{tabular}{|l|}
  \hline
  Typical ConvTranspose Module \\ \hline \hline
  SpatialConvTranspose, 4x4\\ \hline
  SpatialConv, 3x3\\ \hline
  SpatialBatchNorm\\ \hline
  ReLU\\ \hline
  
\end{tabular}

\tikzset{%
  >={Latex[width=2mm,length=2mm]},
            base/.style = {rectangle, rounded corners,draw=black,
                           minimum width=2.2cm},
            Conv/.style = {base, fill=blue!30},
            Deconv/.style = {base, fill=red!30},
            Concat/.style = {base, fill=green!30},
            Depth/.style = {base, fill=orange!15},
            Loss/.style = {base, fill=purple!30},
}

\begin{tikzpicture}[scale = 0.5]
  \node (start)             [base]              {Input image pair};
  \node (Conv1)      [Conv, below of=start]          {Conv1, stride 2};
  \node (Conv2)      [Conv, below of=Conv1]   {Conv2, stride 2};
  \node (Conv3)      [Conv, below of=Conv2]   {Conv3, stride 2};
  \node (Conv3_1)    [Conv, below of=Conv3]   {Conv3.1};
  \node (Conv4)      [Conv, below of=Conv3_1]   {Conv4, stride 2};
  \node (Conv4_1)    [Conv, below of=Conv4]   {Conv4.1};
  \node (Conv5)      [Conv, below of=Conv4_1]   {Conv5, stride 2};
  \node (Conv5_1)    [Conv, below of=Conv5]   {Conv5.1};
  \node (Conv6)      [Conv, below of=Conv5_1]   {Conv6, stride 2};
  \node (Conv6_1)    [Conv, below of=Conv6]   {Conv6.1};
  \node (Deconv5)    [Deconv, below of=Conv6_1] {Deconv5};
  \node (Concat5)    [Concat, below of=Deconv5] {Concat5};
  \node (Deconv4)    [Deconv, below of=Concat5] {Deconv4};
  \node (Concat4)    [Concat, below of=Deconv4] {Concat4};
  \node (Deconv3)    [Deconv, below of=Concat4] {Deconv3};
  \node (Concat3)    [Concat, below of=Deconv3] {Concat3};
  \node (Deconv2)    [Deconv, below of=Concat3] {Deconv2};
  \node (Concat2)    [Concat, below of=Deconv2] {Concat2};
  \node (Depth2)     [Depth, below of=Concat2] {Depth2};
  \node (finish)      [base, below of=Depth2] {Final depth output};
  \node (L1Loss)     [Loss, right of=Depth2, xshift=4.5cm, yshift=-1cm,text width=2.2cm, align=center] {MultiScale L1 Loss};
  
  \node (Depth6)     [Depth, right of=Conv6_1, xshift=2.5cm, yshift= -0.5cm] {Depth6};
  \node (UpDepth6)    [Deconv, below of=Depth6] {Up Depth6};
  
  \node (Depth5)     [Depth, right of=Concat5, xshift=2.5cm, yshift= -0.5cm] {Depth5};
  \node (UpDepth5)    [Deconv, below of=Depth5] {Up Depth5};
  
  \node (Depth4)     [Depth, right of=Concat4, xshift=2.5cm, yshift= -0.5cm] {Depth4};
  \node (UpDepth4)    [Deconv, below of=Depth4] {Up Depth4};
  
  \node (Depth3)     [Depth, right of=Concat3, xshift=2.5cm, yshift= -0.5cm] {Depth3};
  \node (UpDepth3)    [Deconv, below of=Depth3] {Up Depth3};
     
  \draw[->]             (start) -- (Conv1) node [midway,right] {6xHxW};
  \draw[->]     (Conv1) -- (Conv2) node [midway,right] {32x\nicefrac{1}{2}Hx\nicefrac{1}{2}W};
  \draw[->]     (Conv2) -- (Conv3)node [midway,right] {64x\nicefrac{1}{4}Hx\nicefrac{1}{4}W};
  \draw[->]     (Conv3) -- (Conv3_1)node [midway,right] {128x\nicefrac{1}{8}Hx\nicefrac{1}{8}W};
  \draw[->]     (Conv3_1) -- (Conv4);
  \draw[->]     (Conv4) -- (Conv4_1)node [midway,right] {256x\nicefrac{1}{16}Hx\nicefrac{1}{16}W};
  \draw[->]     (Conv4_1) -- (Conv5);
  \draw[->]     (Conv5) -- (Conv5_1)node [midway,right] {256x\nicefrac{1}{32}Hx\nicefrac{1}{32}W};
  \draw[->]     (Conv5_1) -- (Conv6);
  \draw[->]     (Conv6) -- (Conv6_1)node [midway,right] {512x\nicefrac{1}{64}Hx\nicefrac{1}{64}W};
  \draw[->]     (Conv6_1) -- (Deconv5);
  \draw[->]     (Conv5_1.west) -- +(-0.6,0) |- (Concat5);
  \draw[->]     (Deconv5) -- (Concat5)node [midway,right] {256x\nicefrac{1}{32}Hx\nicefrac{1}{32}W};
  \draw[->]     (UpDepth6) |- (Concat5);
  \draw[->]     (Concat5) -- (Deconv4);
  \draw[->]     (Deconv4) -- (Concat4)node [midway,right] {128x\nicefrac{1}{16}Hx\nicefrac{1}{16}W};
  \draw[->]     (Conv4_1.west) -- +(-0.8,0) |- (Concat4);
  \draw[->]     (UpDepth5) |- (Concat4);
  \draw[->]     (Concat4) -- (Deconv3);
  \draw[->]     (UpDepth4) |- (Concat3);
  \draw[->]     (Conv3_1.west) -- +(-1,0) |- (Concat3);
  \draw[->]     (Deconv3) -- (Concat3)node [midway,right] {64x\nicefrac{1}{8}Hx\nicefrac{1}{8}W};
  \draw[->]     (Concat3) -- (Deconv2);
  \draw[->]     (Conv2.west) -- +(-1.2,0) |- (Concat2);
  \draw[->]     (Deconv2) -- (Concat2)node [midway,right] {32x\nicefrac{1}{4}Hx\nicefrac{1}{4}W};
  \draw[->]     (UpDepth3) |- (Concat2);
  \draw[->]     (Concat2) -- (Depth2);
  \draw[->]     (Conv6_1) |- (Depth6);
  \draw[->]     (Concat5) |- (Depth5);
  \draw[->]     (Concat4) |- (Depth4);
  \draw[->]     (Concat3) |- (Depth3);
  \draw[->]     (Depth6) -- (UpDepth6)node [midway,right] {1x\nicefrac{1}{32}Hx\nicefrac{1}{32}W};
  \draw[->]     (Depth5) -- (UpDepth5)node [midway,right] {1x\nicefrac{1}{16}Hx\nicefrac{1}{16}W};
  \draw[->]     (Depth4) -- (UpDepth4)node [midway,right] {1x\nicefrac{1}{16}Hx\nicefrac{1}{16}W};
  \draw[->]     (Depth3) -- (UpDepth3)node [midway,right] {1x\nicefrac{1}{8}Hx\nicefrac{1}{8}W};
  \draw[->]     (Depth6) -| (L1Loss);
  \draw[->]     (Depth5) -| (L1Loss);
  \draw[->]     (Depth4) -| (L1Loss);
  \draw[->]     (Depth3) -| (L1Loss);
  \draw[->]     (Depth2) -| (L1Loss);
  \draw[->]     (Depth2) -- (finish) node [midway,right] {1x\nicefrac{1}{4}Hx\nicefrac{1}{4}W};
  \end{tikzpicture}
\linebreak
\caption{DepthNet structure parameters}
\label{network}
\end{figure}
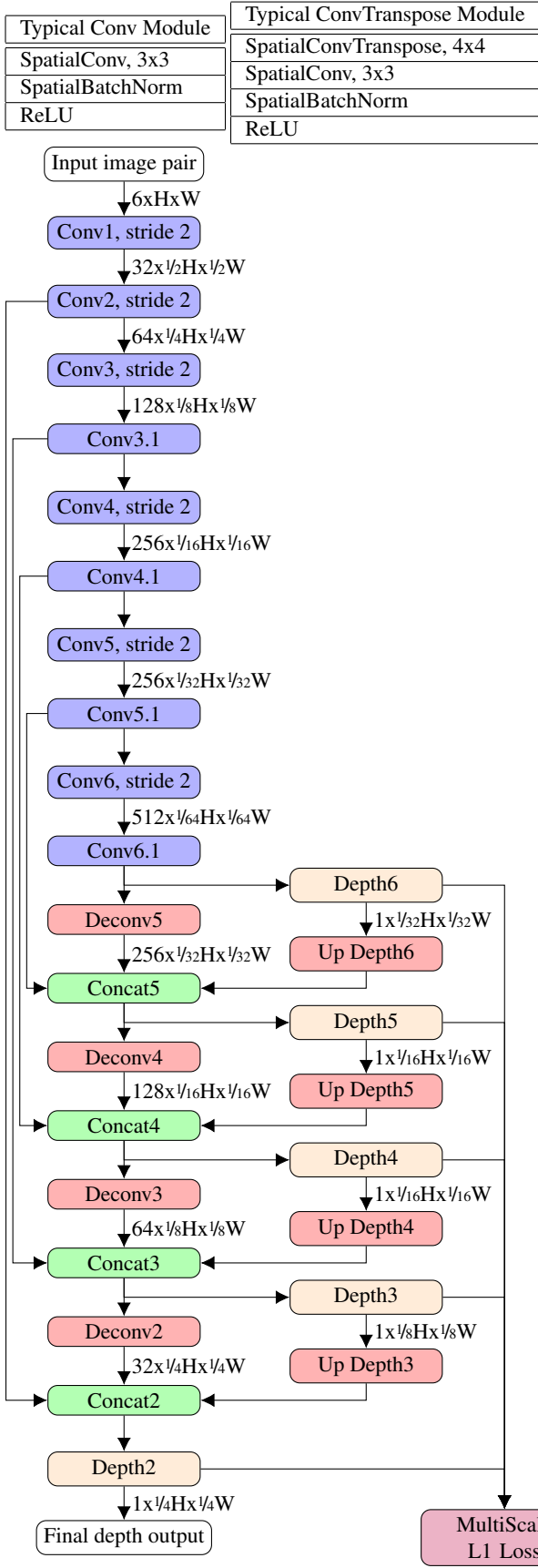

Our network, which is broadly inspired from FlowNetS \cite{DFIB15} and called DepthNet is described Fig~\ref{network}. This network was initially used for flow inference. The main idea behind this network is that upsampled feature maps are concatenated with corresponding earlier convolution outputs. Higher semantic information is then associated with information more closely linked to pixels (since it went through less strided convolutions) which is then used for reconstruction.

This has been proven very efficient for flow and disparity computing while keeping a very simple supervised learning process. The architecture is admittedly very simple and one could leverage some more advanced work for flow and disparity, such as FlowNetC or GC-Net \cite{2017arXiv170304309K} among many others.
The main point of this experimentation is to show that direct depth estimation can be beneficial regarding unknown translation. Like FlowNetS, we use a multi-scale criterion, with a L1 reconstruction error for each scale.

\begin{equation}
Loss = \sum_{s\in scales} \gamma_s\frac{1}{H_sW_s} \sum_i \sum_j \left| output_s(i,j) - depth_s(i,j)\right|
\end{equation}
where
\begin{itemize}
\setlength\itemsep{0em}\setlength\parskip{0em}\setlength\topsep{0em}\setlength\partopsep{0em}\setlength\parsep{0em}
\item $\gamma_s$ is the weight of the scale, arbitrarily chosen as $W_s$ in our experiments.
\item $(H_s,W_s) = (\nicefrac{1}{2^n}H,\nicefrac{1}{2^n}W)$ are the height and width of the output.
\item $depth_s$ is the scaled depth groundtruth, using average pooling.
\end{itemize}

As said earlier, we apply data augmentation to the dataset using different shifts, along with classic methods such a flips and rotations. We also clamp depth to a maximum of 100m, and provide sample pair without shift, assuming its depth is 100m everywhere.

\begin{figure}
\includegraphics[scale=0.4]{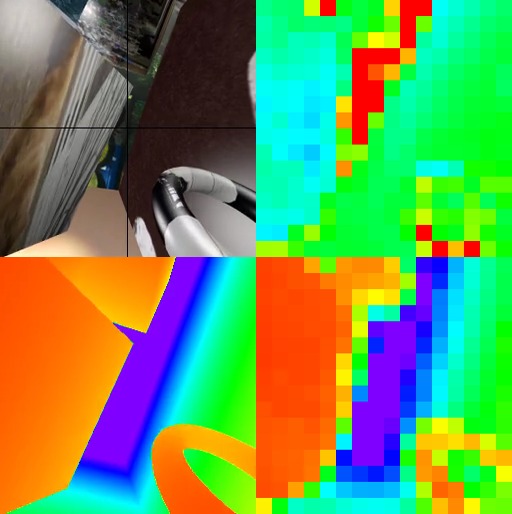}
\caption{result for 64x64 images, upper-let : input (before being downscaled to 64x64), lower-left : Ground Truth depth, lower-right : our network output (16x16), upper-right : error, green is no error, red is overestimated depth, blue is sub estimated}
\label{result64}
\end{figure}

\begin{figure}
\includegraphics[scale=0.2]{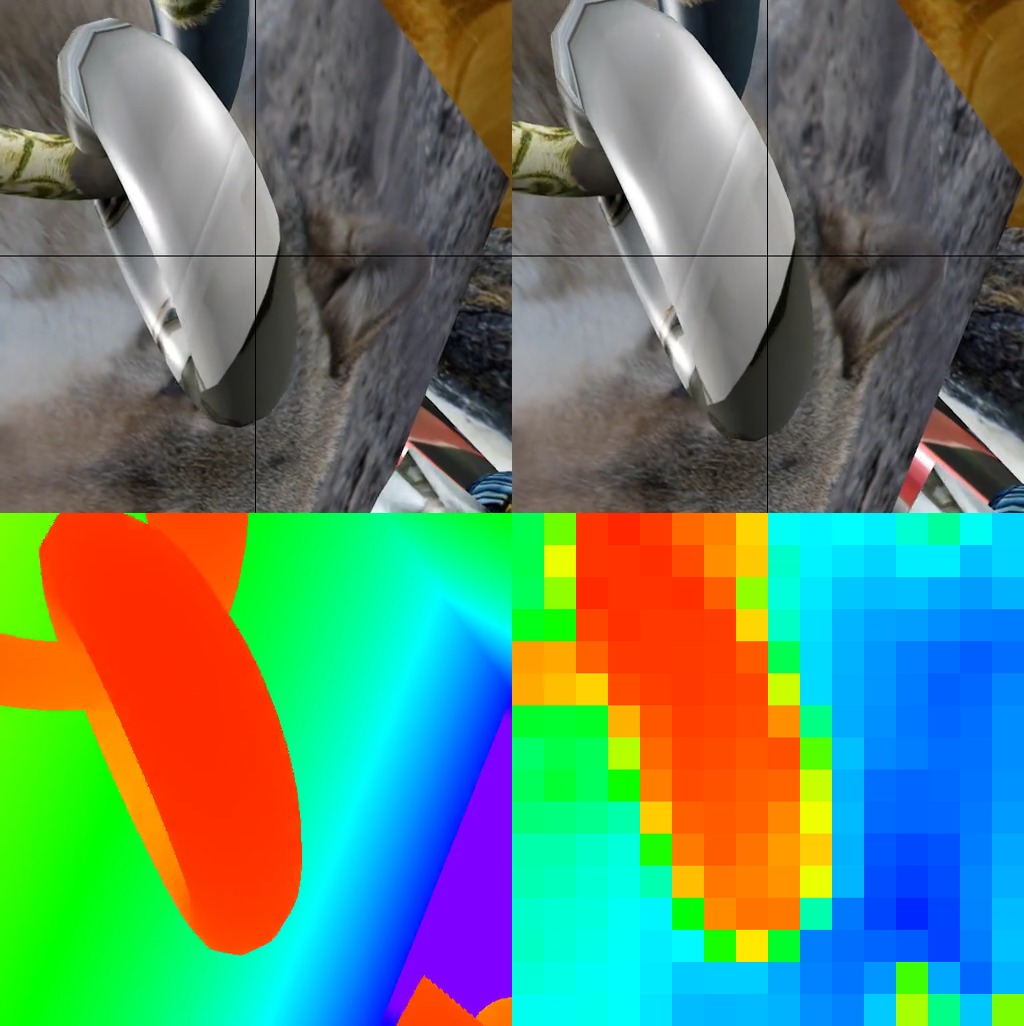}
\caption{(On top, $img_t$ and $img_{t+3}$) Result for forward movement, showing that the network is also doing shape identification}
\label{forward}
\end{figure}

Fig~\ref{result64} shows results from 64px dataset. Like FlowNetS, results are downsampled by a factor of 4, which gives 16x16 Depth Maps. 

One can notice that although the network is still fully convolutional, feature map sizes go down to 1x1 and then behave exactly like a Fully Connected Layer, which can serve to figure out implicitly motion direction and spread this information across the outputs. 
The second noticeable fact is that near FOE, (see Fig~\ref{forward} for centered FOE, i.e. perfect forward movement) the network has no problem inferring depth, which means that it uses neighbor disparity and interpolates when no other information is available.

This can be interpreted as 3d shapes identification, along with their magnification : pixels belonging to the same shape are deemed to have close and continuous depth values, resulting in a FOE-independent depth inference.

\subsection{From 64px to 512px Depth inference}

\begin{figure}
\begin{tabular}{l}
  \includegraphics[scale=0.2]{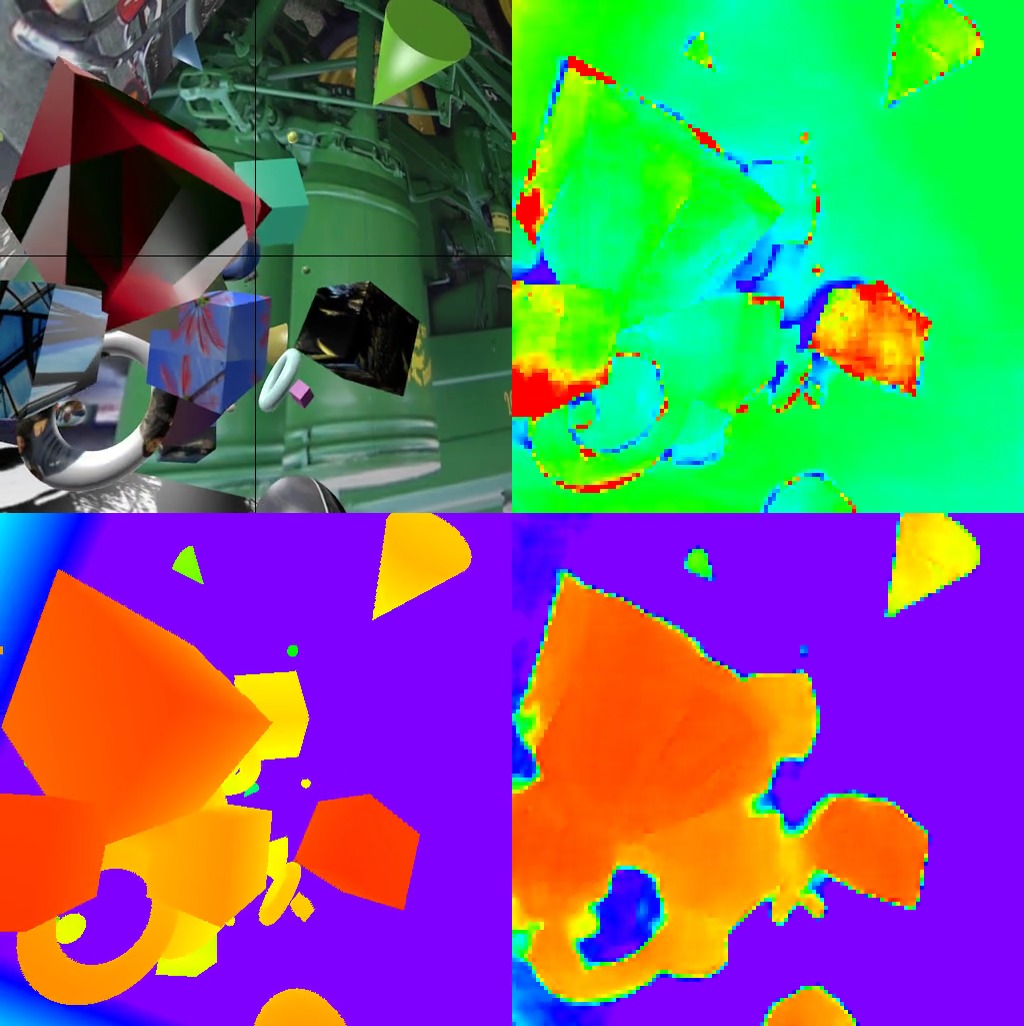} \\ \includegraphics[scale=0.2]{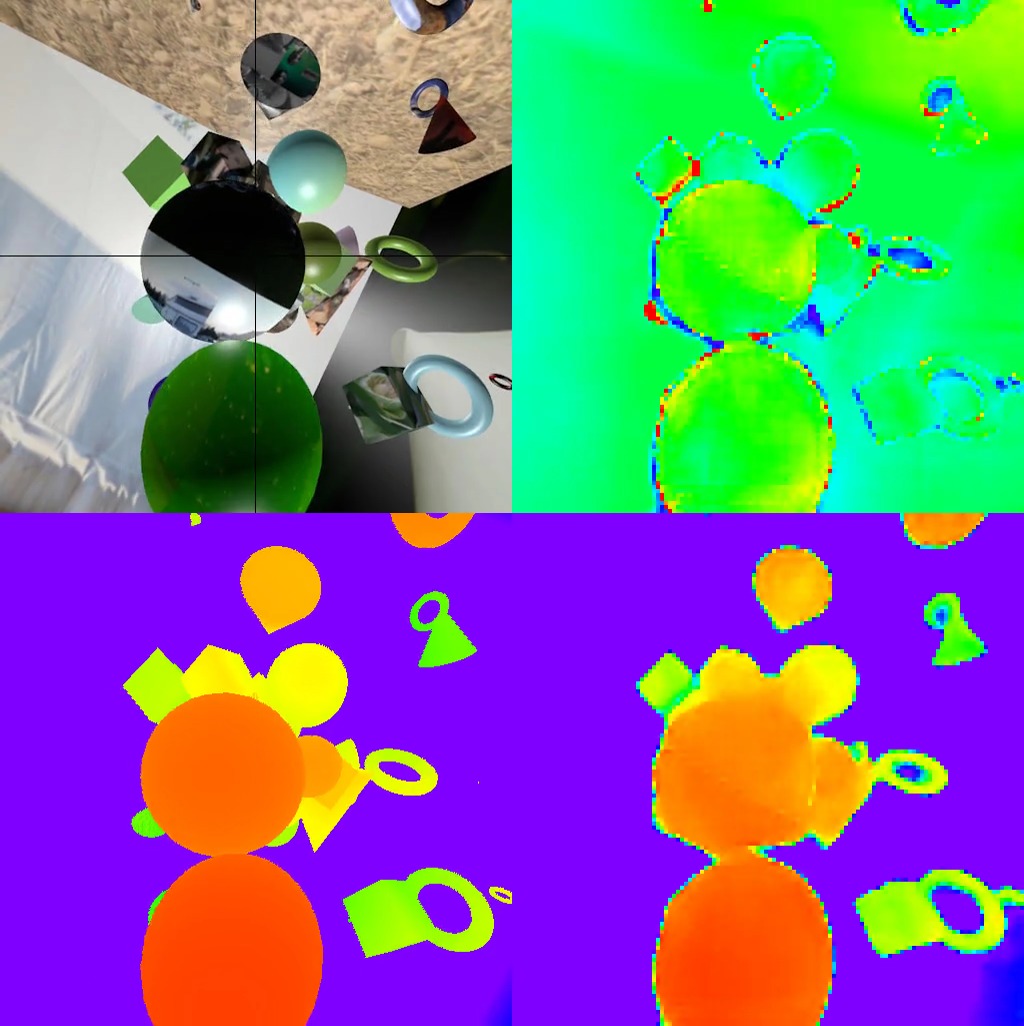}
\end{tabular}
\caption{some results on 512x512 images, same color code as for 64x64 input}
\label{result512}
\end{figure}

\begin{figure}
\begin{tabular}{l}
  \includegraphics[scale=0.22]{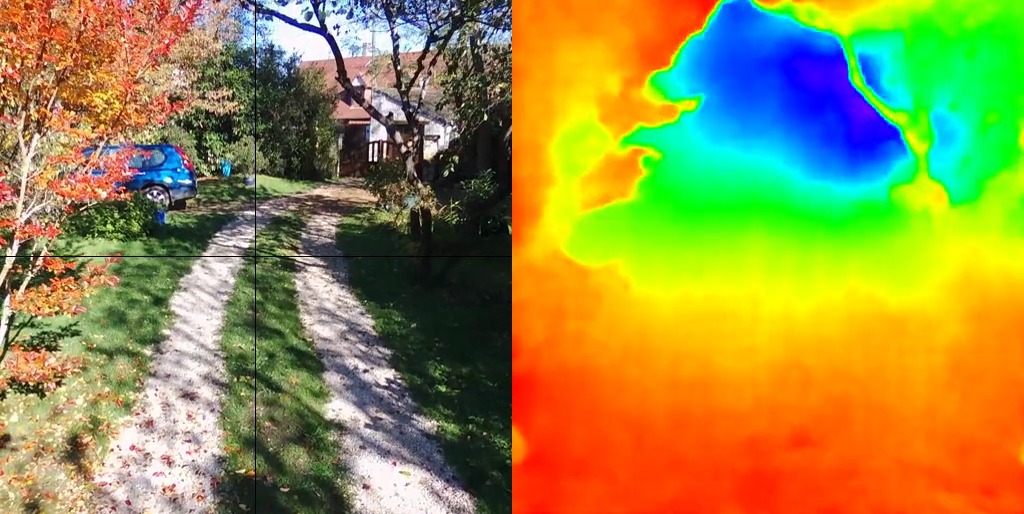} \\ \includegraphics[scale=0.22]{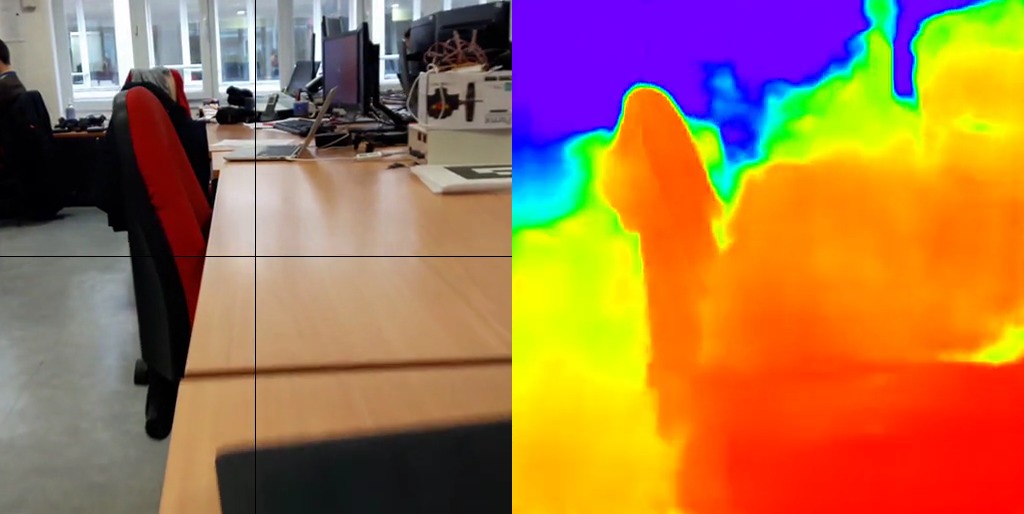}
\end{tabular}
\caption{some results on real images input. Up is from a Bebop drone footage, down is from a gimbal stabilized smartphone video}
\label{resultirl}
\end{figure}

\begin{table}
\centering
\begin{tabular}{|l|c|c|c|c|c}
  \hline
  \multirow{2}{*}{Network} & \multicolumn{2}{|c|}{L1Error} & \multicolumn{2}{|c|}{RMSE}\\ \cline{2-5}
  &train&test&train&test\\ \hline
  FlowNetS$_{64}$&$1.69$&$4.16$&$4.25$&$7.97$\\ \hline
  DepthNet$_{64}$&$2.26$&$4.49$&$5.55$&$8.44$		\\ \hline
  FlowNetS$_{64\rightarrow128\rightarrow256\rightarrow512}$&$0.658$&$\textbf{2.44}$&$1.99$&$\textbf{4.77}$\\ \hline
  DepthNet$_{64\rightarrow128}$&$1.20$&$3.07$&$3.43$&$6.30$\\ \hline
  DepthNet$_{64\rightarrow128\rightarrow256}$&$0.876$&$\textbf{2.44}$&$2.69$&$4.99$\\ \hline
  DepthNet$_{64\rightarrow128\rightarrow256\rightarrow512}$&$1.09$&$2.48$&$2.86$&$\textbf{4.90}$\\ \hline
  DepthNet$_{64\rightarrow512}$&$1.02$&$2.57$&$2.81$&$5.13$\\ \hline
  DepthNet$_{512}$&$1.74$&$4.59$&$4.91$&$8.62$\\ \hline
  
\end{tabular}
\caption{quantitative results for depth inference networks. FlowNetS is modified with 1 channel outputs (instead of 2 for flow), trained from scratch for depth with Still Box}
\label{quantitative}
\end{table}
\begin{table}
\centering
\begin{tabular}{|l|c|c|c|c|c|c|c|}
  \hline
  \multirow{2}{*}{Network} & \multirow{2}{*}{size} & \multicolumn{2}{|c|}{980Ti} &  \multicolumn{2}{|c|}{\begin{tabular}{@{}c@{}}Quadro \\ K2200m\end{tabular}}& \multicolumn{2}{|c|}{TX1}\\ \cline{3-8}
  &&$1$&$8$&$1$&$8$&$1$&$8$\\ \hline
  FlowNetS$_{64}$&$39.4$&$225$&$153$&$76$&$41$&$29$&$14$\\ \hline
  DepthNet$_{64}$&$\textbf{7.33}$&$364$&$245$&$190$&$124$&$70$&$40$		\\ \hline
  FlowNetS$_{512}$&$39.4$&$69$&$8.8$&$16$&N/A&$2.8$&N/A\\ \hline
  DepthNet$_{128}$&$\textbf{7.33}$&$294$&$118$&$171$&$75$&$51$&$15$\\ \hline
  DepthNet$_{256}$&$\textbf{7.33}$&$178$&$36$&$121$&$30$&$39$&$3.2$\\ \hline
  DepthNet$_{512}$&$\textbf{7.33}$&$68$&$8.8$&$51$&$7.6$&$9.2$&N/A\\ \hline
\end{tabular}
\caption{Size (millions of parameters) and Inference speeds (fps) on different devices. Batch sizes are $1$ and $8$ (when applicable). A batch size of $8$ means $8$ depth maps are computed at the same time}
\label{speed}
\end{table}

One could think that a fully convolutional network such as ours can not solve depth extraction for pictures greater than 64x64. The main idea is that for a fully convolutional network, each pixel is applied the same operation. For disparity, this makes sense because the problem is essentially similarity from different picture shifts.
Wherever we are on the picture, the operation is the same. For depth inference when FOE is not diverging (forward movement is non negligible), result from Theorem~\ref{D2D} apparently shows that once you know the FOE, you then get different operations to do depending on your distance from it and from the optical center $\textbf{P}_0$.
The only possible strategy for a fully convolutional network would be to compute the position in the frame as well and to apply the compensating scaling to the output.

This problem then seems very difficult, if not impossible for a network as simple as ours, and if we run the training directly on 512x512 images, the network fails to converge to better results than with 64x64 images (while better resolution would help getting more precision). However, if we take the converged network and apply a fine-tuning on it with 512x512 images, we get much better results. Fig~\ref{result512} shows training results for mean L1 reconstruction error, and shows that our deemed-impossible problem seems to be easily solved with multi-scale fine-tuning. As Table~\ref{quantitative} shows, best results are obtained with multiple fine-tuning, with intermediate scales $64$, $128$, $256$,  and finally $512$ pixels. Subscript values indicate finetuning processes. FlowNetS is performing better than DepthNet but by a fairly light margin while being 5 times heavier and most of the time much slower, as shown Table~\ref{speed}.

Fig~\ref{resultirl} shows qualitative results from our validation set, and from real condition drone footage, on which we were careful to avoid camera rotation. These results did not benefit from any fine-tuning from real footage, indicating that our Still Box Dataset, although not realistic in its scenes structures and rendering, appears to be sufficient for learning to produce decent depthmaps in real conditions.

\subsection{Quality measurement}
As our network is leveraging the reduced dimensionality of our dataset due to its lack of rotation, it is hard to compare our method to anything else. Disparity estimation is equivalent to a lateral translation that our network has been trained on, and could be used to compare to other algorithms but this reduced context seems unfair compared to methods designed especially for it.

Other datasets provide ego motion with 6-DOF on which our network has not been trained and is certain to give poor results. On the other hand, we could test some SLAM methods but they work better when applied to long image sequences and not only image pairs.
In short, our method is setting state of the art, but for a very particular problem that we hope will gain interest with time.

\section{UAV navigation use-case}
We assumed in learning depth inference from a moving camera, assuming its velocity is always the same. When running during flight, such a system can easily deduce the real depth map from the drone speed $V_t$, knowing that the training speed was $V_0$ (here $9m.s^{-1}$)
\begin{equation}
depth(t) = \frac{V_t}{V_0} DepthNet(frame_t,frame_{t-1})
\end{equation}

One of the drawbacks of this learning method is that the $f$ value (which is focal length divided by sensor size per pixel) of our camera must be the same as the one used in training. Our dataset creation framework however allows us to change this value very easily for training. One must also be sure to have pinhole equivalent frames like during training.

\subsection{Multiple shifts inference}

Depending of the depth distribution of the groundtruth depth map, it may be useful to adjust frame shift. For example, when flying high above the ground, big structure detection and avoidance requires knowing precise distance values that are outside the typical range of any RGB-D sensor. The logical strategy would then be to increase the temporal shift between the frame pairs provided to DepthNet as inputs.

More generally, one must ensure a well distributed depth map from 0 to 100m to get high quality depth inference. This problem can be solved with two (among other) solutions:
\begin{itemize}
\item Deduce optimal shift $\Delta_t$ from precedent inference distribution, e.g:
\[
\Delta_{t+1} = \Delta_t\frac{E_{depth}}{E_0}
\]
where $E_0$ is 50m (because our network outputs from 0 to 100m) and $E_{depth}$ is the mean of precedent output, i.e. :
\[
E_{depth} = \frac{1}{HW}\sum_{i,j} DepthNet(frame_t,frame_{t-\Delta_t})_{i,j}
\]
\item Use batch inference to compute depth with multiple shifts $\Delta_{t,i}$. As shown in Table~\ref{speed}, batch size greater than 1 can be used to some extent (especially for low resolution) to efficiently compute multiple depth maps.
\[
Depth_i(t) = DepthNet(frame_t,frame_{t-\Delta_{i,t}})
\]
These multiple depth maps can then be either combined to construct a high quality depth map, or used separately to run two different obstacle avoidance algorithm, e.g. one dedicated for long range path planning (and then a high value $\Delta_{i,t}$) and the other for reactive and short range collision avoidance with low $\Delta_{i,t}$.
While one depth map will display closer areas at zero distance but further regions with precision, the other will set far regions to infinity (or 100m for DepthNet) but closer region with high resolution as flow is lowered compared to a high shift, and potentially within the range the network has been trained on.
\end{itemize}
\pagebreak
\section{Conclusion and future work}
We propose a novel way of computing dense depth maps from motion, along with a very comprehensive dataset for stabilized footage analysis. This algorithm can then be used for depth-based sense and avoid algorithm in a very flexible way, in order to cover all kinds of path planning, from collision avoidance to long range obstacle bypassing.

Future works include implementation of such a path planning algorithm, and construction of a real condition fine tuning dataset, using UAVs footages and a preliminary thorough 3D offline scan. This would allow us to measure quantitative quality of our network for real footages and not only subjective as for now.

We also believe that our network can be extended to reinforcement learning applications that will potentially result in a complete end-to-end sense and avoid neural network for monocular cameras.

\bibliography{biblio}

\appendix
\section{Appendix A : Proof of Theorem~\ref{D2D}}

For a random rotation-less displacement of norm $D$, depth is an explicit function of disparity, focus of expansion $\textbf{FOE}$ and optical center $\textbf{P}_0$

\[
\forall \textbf{P},depth(\textbf{P}) = \frac{Vf}{\sqrt{f^2 + \left\lVert \textbf{P}_0 - \textbf{FOE} \right\rVert^2}} \left(\frac{\left\lVert \textbf{P} - \textbf{FOE} \right\rVert}{disparity(\textbf{P})}-1\right)
\]

\begin{Proof}
We assume no rotation. which means $\textbf{FOE}$ is projection of B on A.

\begin{tikzpicture}[scale=2]
    \draw [->] (0,0) -- +(0.3,0.7) node [midway,left] {$\textbf{m}$};
    \draw (0,0)node [left] {$A$} -- +(-1.5,2);
    \draw (0,0) -- +(1.5,2);
    \draw (0.3,0.7) node[right]{$B$}-- +(-0.97,1.3);
    \draw (0.3,0.7) -- +(0.97,1.3);
    \draw [dashed] (0,0) -- +(0,2);
    \draw [dashed] (0.3,0.7) -- +(0,1.3);
\end{tikzpicture}

Let $\textbf{m}$ = $\begin{pmatrix}m_x\\m_y\\m_z
\end{pmatrix}$ be $\overrightarrow{AB}$

\begin{equation}
\label{FOE}
\textbf{FOE} = \begin{pmatrix}
FOE_u\\
FOE_v
\end{pmatrix} = \begin{pmatrix}
u_0 + f\frac{m_x}{m_z}\\
v_0 + f\frac{m_y}{m_z}
\end{pmatrix}
\end{equation}

let $\textbf{P}_{XYZ}$ be a point $\begin{pmatrix}
X\\Y\\Z
\end{pmatrix}$. For camera B we have $\textbf{P}_B = \begin{pmatrix}u_B\\v_B\end{pmatrix}
= \begin{pmatrix}u_0 + f\frac{X}{Z}\\v_0 + f \frac{Y}{Z}\end{pmatrix}$

relative movement of $\textbf{P}_{XYZ}$ is $-\textbf{m}$

so we have $\left\lbrace\begin{array}{c}
dX = -m_x\\
dY = -m_y\\
dZ = -m_z
\end{array}\right.
$

If we compute $u_A$ for $\textbf{P}_A$ : 
\[u_A = u_0 + f\frac{X-dX}{Z-dZ}\]
\[
du = u_B - u_A = f\left(\frac{X}{Z} - \frac{X+m_x}{Z+m_z}\right)
\]
\[du = \frac{f}{Z+m_z}\left(-m_x +\frac{X}{Z}m_z\right)
\]
\[
du = \frac{m_z}{Z+m_z}\left(u_B - FOE_u\right)
\]

Similarly, with $v$, we get:

\begin{equation}
\left\lbrace \begin{array}{c}
du = \frac{m_z}{Z + m_z}\left(u_B-FOE_u\right)\\
dv = \frac{m_z}{Z + m_z}\left(v_B-FOE_v\right)
\end{array}\right.
\end{equation}

We consider disparity as norm of the flow $\begin{pmatrix}
du\\dv
\end{pmatrix}$ expressed in frame B (which is correlated to depth at this frame).

\begin{equation}
\forall \textbf{P} = \begin{pmatrix}
u\\v\end{pmatrix},disparity(\textbf{P}) = \left\lVert\begin{array}{c}
du\\dv
\end{array}\right\rVert = \frac{m_z}{Z+m_z} \left\lVert \textbf{P}- \textbf{FOE} \right\rVert
\end{equation}

Consequently, we can deduce depth at frame B from disparity :

\begin{equation}
\label{depth}
\forall \textbf{P} = \begin{pmatrix}
u\\v\end{pmatrix},depth(\textbf{P}) = m_z\left(\frac{\left\lVert \textbf{P} - \textbf{FOE}\right\rVert}{disparity(\textbf{P})}-1\right)
\end{equation}

From our dataset construction, we know that $m_x^2~+~m_y^2~+~m_z^2~=~V^2$
Let us call $\left\lbrace\begin{array}{l}
\Delta u = u_0 - FOE_u \\
\Delta v = v_0 - FOE_v
\end{array}
\right.
$

From \ref{FOE}, we get:
\[
\left\lbrace\begin{array}{l}
m_x = \frac{m_z \Delta u}{f} \\
m_y = \frac{m_z \Delta v}{f}
\end{array}
\right.
\]

\[
V^2 = m_z^2\left(1+\frac{\Delta u^2 + \Delta v^2}{f^2}\right)
\]

\[
m_z = \frac{Vf}{\sqrt{f^2 + \Delta u^2 + \Delta v^2}}
\]
and then from \ref{depth} we get

\begin{equation}
\forall \textbf{P},depth(\textbf{P}) = \frac{Vf}{\sqrt{f^2 + \left\lVert \textbf{P}_0 - \textbf{FOE} \right\rVert^2}} \left(\frac{\left\lVert \textbf{P} - \textbf{FOE} \right\rVert}{disparity(\textbf{P})}-1\right) \square
\end{equation}

\end{Proof}

\end{document}